\documentclass[conference]{IEEEtran}
\IEEEoverridecommandlockouts
\usepackage{cite}
\usepackage{amsmath,amssymb,amsfonts}
\usepackage{algorithmic}
\usepackage{graphicx}
\usepackage{subcaption}
\usepackage{textcomp}
\usepackage{xcolor}
\usepackage{enumitem}
\usepackage{tikz}
\usepackage{pifont}
\usepackage{url}
\usepackage{color}
\usepackage{colortbl} 
\usepackage{xcolor} 
\usepackage{fancyhdr}
\usepackage{marvosym}
\usepackage{diagbox}

\usepackage{
rotating,
booktabs,
mathrsfs,
url,
etoolbox,
multirow,
hyperref,
manyfoot,
hhline,
pifont,
cleveref}
\newcommand\shline{\specialrule{0.8pt}{0pt}{0pt}}
\hypersetup{
colorlinks=true,
linkcolor=black
}
\definecolor{darkgreen}{rgb}{0, 0.7, 0}
\definecolor{darkred}{rgb}{0.8, 0, 0}
\newcommand*\colorcheck{%
  \expandafter\newcommand\csname greencheck\endcsname{\textcolor{darkgreen}{\ding{52}}}%
}
\newcommand*\colorcross{%
  \expandafter\newcommand\csname redcross\endcsname{\textcolor{darkred}{\ding{56}}}%
}
\colorcheck
\colorcross
\definecolor{skyblue}{rgb}{0.16, 0.65, 0.87}
\definecolor{orange}{rgb}{0.93, 0.49, 0.19}
\definecolor{green}{rgb}{0.44, 0.68, 0.28}
\definecolor{navy}{rgb}{0.27, 0.45, 0.78}
\def\BibTeX{{\rm B\kern-.05em{\sc i\kern-.025em b}\kern-.08em
    T\kern-.1667em\lower.7ex\hbox{E}\kern-.125emX}}
\begin{document}

\title{TalkFashion: Intelligent Virtual Try-On Assistant Based on Multimodal Large Language Model}


\author{
Yujie Hu$^{1}$,
Xuanyu Zhang$^{1}$,
Weiqi Li$^{1}$,
Jian Zhang$^{1,2\dag}$
\\
\emph{$^{1}$School of Electronic and Computer Engineering, Peking University, China}\\
\emph{$^{2}$Guangdong Provincial Key Laboratory of Ultra High Definition Immersive Media Technology,}\\
\emph{Shenzhen Graduate School, Peking University}
}

\maketitle
\let\thefootnote\relax\footnotetext{$^\dag$Corresponding author: Jian Zhang (zhangjian.sz@pku.edu.cn). This work was supported in part by National Natural Science Foundation of China (No. 62372016), Guangdong Provincial Key Laboratory of Ultra High Definition Immersive Media Technology (No. 2024B1212010006).}


\begin{abstract}
Virtual try-on has made significant progress in recent years. This paper addresses how to achieve multifunctional virtual try-on guided solely by text instructions, including full outfit change and local editing. Previous methods primarily relied on end-to-end networks to perform single try-on tasks, lacking versatility and flexibility. We propose TalkFashion, an intelligent try-on assistant that leverages the powerful comprehension capabilities of large language models to analyze user instructions and determine which task to execute, thereby activating different processing pipelines accordingly. Additionally, we introduce an instruction-based local repainting model that eliminates the need for users to manually provide masks. With the help of multi-modal models, this approach achieves fully automated local editings, enhancing the flexibility of editing tasks. The experimental results demonstrate better semantic consistency and visual quality compared to the current methods.

\end{abstract}


\section{Introduction}
\label{sec:intro}

Virtual Try-On (VTON) aims to generate a realistic image of a person wearing a specified garment based on input images of the clothing and the individual, typically using in-shop garment images. Due to advancements in generative models, VTON has garnered significant attention and led to the development of multiple application scenarios, such as in-the-wild try-on (transferring garments worn by another person to the target individual) \cite{streettryon, catvton}, multiple-garment compositional try-on (allowing users to freely choose and combine a variety of tops, bottoms, and accessories) \cite{mmvto, mmtryon}, and video virtual try-on (applying a clothing item onto a video of a target individual to display it from various angles) \cite{tunneltryon, vivid}.


On one hand, these tasks are predominantly accomplished by training an end-to-end network on paired datasets. It can automate the processing for a specific task with a single trained network, leading to good results in changing outfits when given a sufficiently rich clothing dataset. However, the drawback lies in its inability to handle multiple tasks simultaneously. Different tasks require distinct datasets with varying characteristics, and mixed training within the same framework can negatively impact performance. On the other hand, the latest diffusion-based methods for virtual try-on treat the task as an inpainting problem, where the diffusion model paints over the area to be changed and then injects the garment information to generate a complete piece of clothing that matches the reference image. This approach does not consider the user's finer-grained needs, such as modifying only a part of a garment while preserving the overall style and design.

\begin{figure}[!t]
  \centering
  \includegraphics[width=1\linewidth]{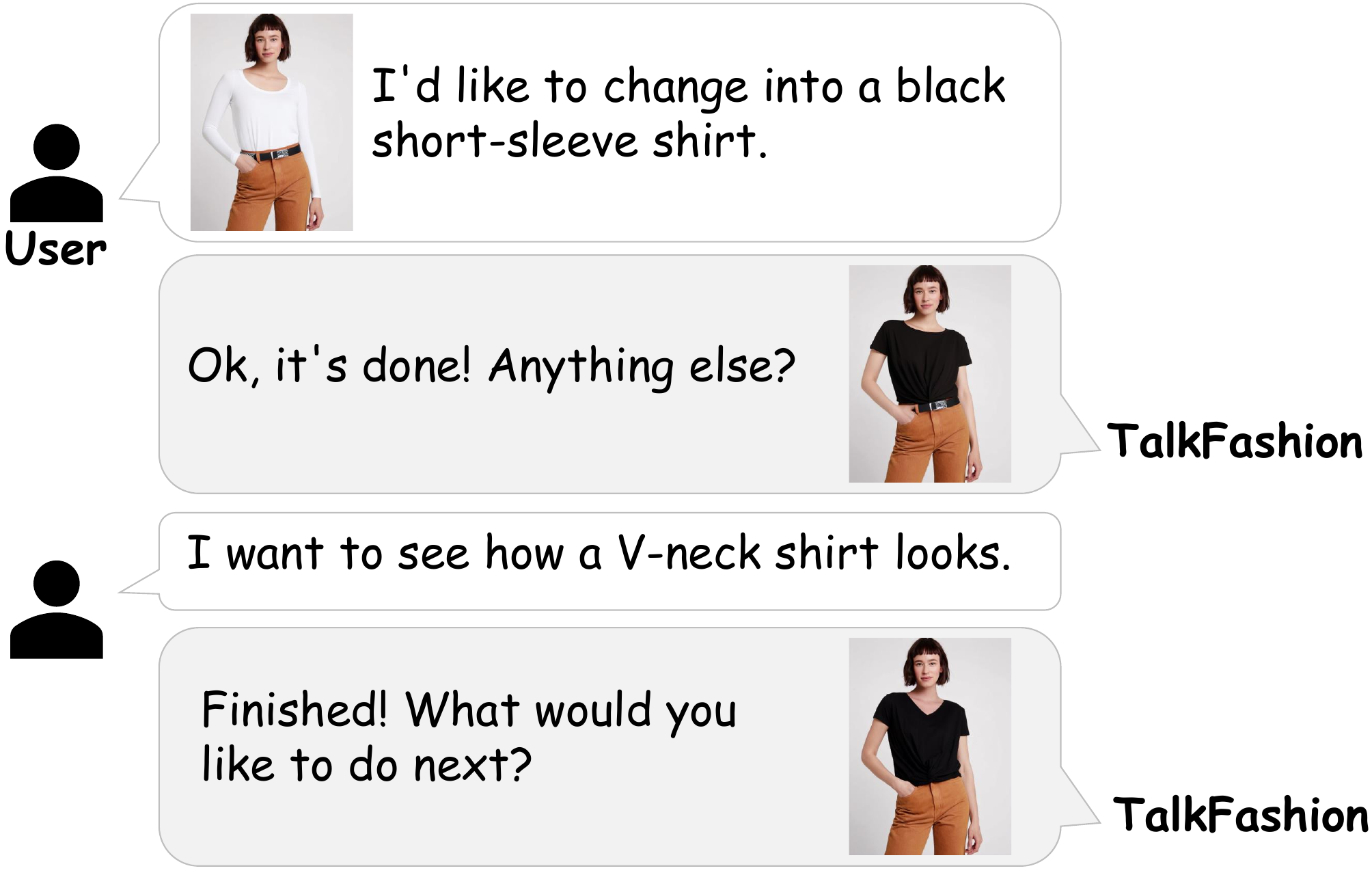}
  \vspace{-14pt}
  \caption{\textbf{Proposed intelligent virtual try-on assistant named TalkFashion.} Users can achieve various outfits by simply inputting text instructions.}
  \vspace{-14pt}
\label{fig-cover} 
\end{figure}

To address these issues and enhance user interaction and flexibility, we propose TalkFashion, an intelligent assistant that integrates multiple virtual try-on functions under the guidance of user text instructions. As shown in Fig. \ref{fig-cover}, TalkFashion supports the combination of different virtual try-on functionalities, including image-based virtual try-on, text-based virtual try-on, and our proposed local editing algorithm. To enable localized editings, we have annotated a mask dataset, utilized a large multimodal model to achieve text-guided localization of local regions, and introduced MLLMs to improve the inpainting network for more realistic local detail generation.

\begin{figure*}[!t]
  \centering
  \includegraphics[width=1\linewidth]{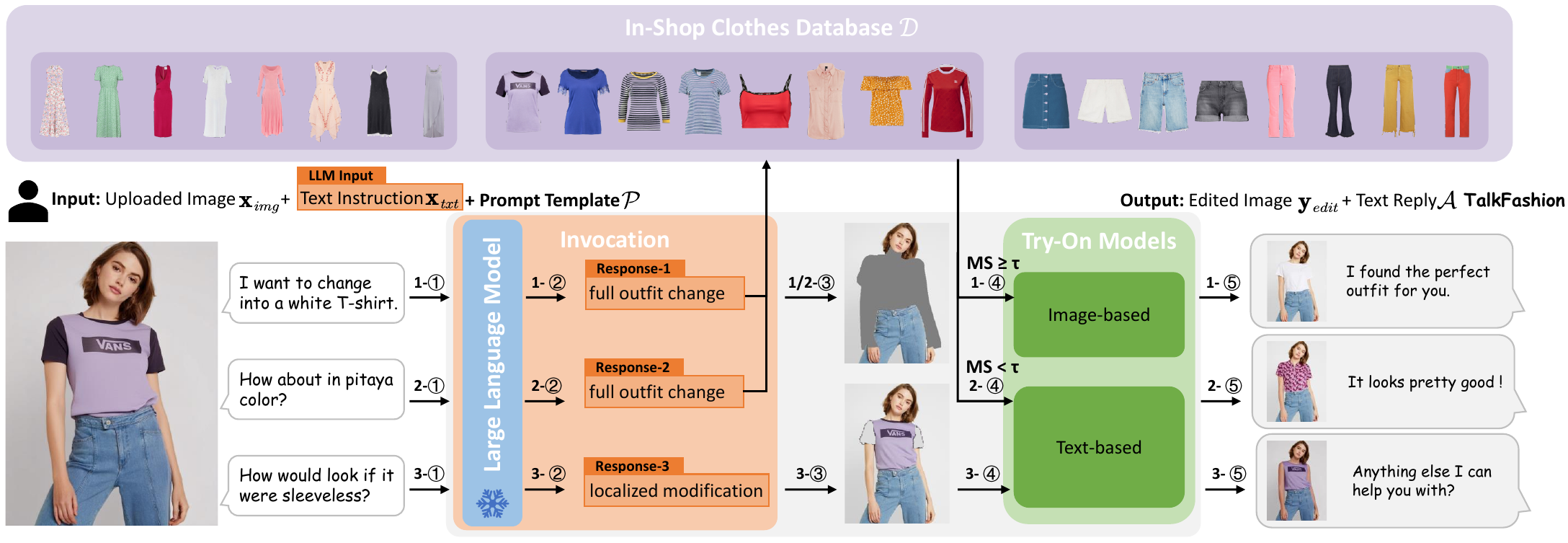}
  \vspace{-14pt}
  \caption{\textbf{Overview of the TalkFashion framework.} We first combine user-input text instructions $\mathbf{x}_{txt}$ with a prompt template $\mathcal{P}$ and input into the LLM\ding{192} to generate structured responses\ding{193}, detailing the functions to be used along with other useful information. Then segment the area to be modified according to the selected function $\mathcal{F}$\ding{194}. If the selected function is full outfit change, the user's instruction is matched with items in the in-shop clothes database $\mathcal{D}$ first. If the match score (MS) is above a predefined threshold $\tau$, the image-based try-on model $\mathcal{G}_{img}$ is used. Otherwise, the text-based try-on model $\mathcal{G}_{txt}$ is applied\ding{195}. Finally, the output image $\mathbf{y}_{edit}$ and the reply $\mathcal{A}$ are returned to the user\ding{196}.}
  \vspace{-14pt}
\label{fig-overview} 
\end{figure*}

In summary, our contributions are as follows:
\begin{itemize}
  \item We propose TalkFashion, an intelligent virtual try-on assistant that allows both full outfit changes and local detail adjustments through simple user text instructions.
  \item We introduce the text-image matching mechanism in the function invocation process, enabling the unification of both image-based and text-based virtual try-on models using pure text instructions for full outfit change.
  \item We propose an instruction-based local repainting model that enables fully automated local editings with text instructions. With the help of multi-modal models, it eliminates the need for users to manually provide masks and achieves more natural and accurate editing results.
\end{itemize}

\section{Related Work}

\subsection{Virtual Try-on}

Current virtual try-on primarily relies on Generative Adversarial Networks (GANs)\cite{gan} or Latent Diffusion Models (LDMs) \cite{stablediffusion} for image synthesis. In GAN-based VTON methods \cite{viton, acgpn, hr-viton, gp-vton}, the main approach involves warping the clothing image and then aligning it with the human's image. However, these methods often struggle with alignment issues caused by explicit flow estimation or distortions and lack generalization to arbitrary person images, such as those with complex backgrounds or intricate poses. With the rapid development of diffusion models \cite{adapter, dragondiffusion, diffeditor}, diffusion model-based virtual try-on methods \cite{stableviton, ootdiffusion, idmvton} have begun to shine. They consider virtual try-on as an exemplar-based image inpainting problem, fine-tuning the inpainting diffusion models on a virtual try-on dataset to generate high-quality virtual try-on images. This approach allows for controlling the generation of clothing through outline sketches, line drawings, or text descriptions, without requiring users to upload specific images of the desired garments. However, these methods still do not fully exploit the semantic information of clothing images, resulting in incomplete preservation of garment details. They also lack the flexibility to perform editings in specific regions, limiting their adaptability to more nuanced user requirements.

\subsection{Tool-augmented Language Models}

Large Language Models (LLMs) \cite{chatgpt, 1vicuna, qwen2, llama, editguard} have made great progress in recent years and have stimulated research in prompt learning and instruction learning. Recent studies \cite{visualchatgpt, gpt4tools, hugginggpt, gorilla} have revealed that LLMs exhibit exceptional capability in invoking various tools, serving as powerful agents to call different tools for target tasks. These models either fine-tune with a dataset of tool-related instructions or translate models or APIs into the language via detailed prompt engineering. The fine-tuning method depends on an instruction-following dataset that includes a wide range of user prompt pairs, yet it is constrained to gathering data for individual and specific tasks only. In terms of prompt engineering, a prevalent approach is to employ a Function Calling mechanism combined with ReAct \cite{react} (Reason+Act). This entails equipping an LLM with a predefined set of functions and user queries, enabling the LLM to dynamically determine which functions to invoke and provide the necessary input parameters for those functions. The output generated by these functions is then returned to the LLM. This approach typically requires sophisticated prompt engineering, but the advantage is that functionality integration can be achieved in a plug-and-play manner by managing prompt templates and function descriptions effectively.

\begin{figure*}[!t]
    \centering
    \includegraphics[width=1\linewidth]{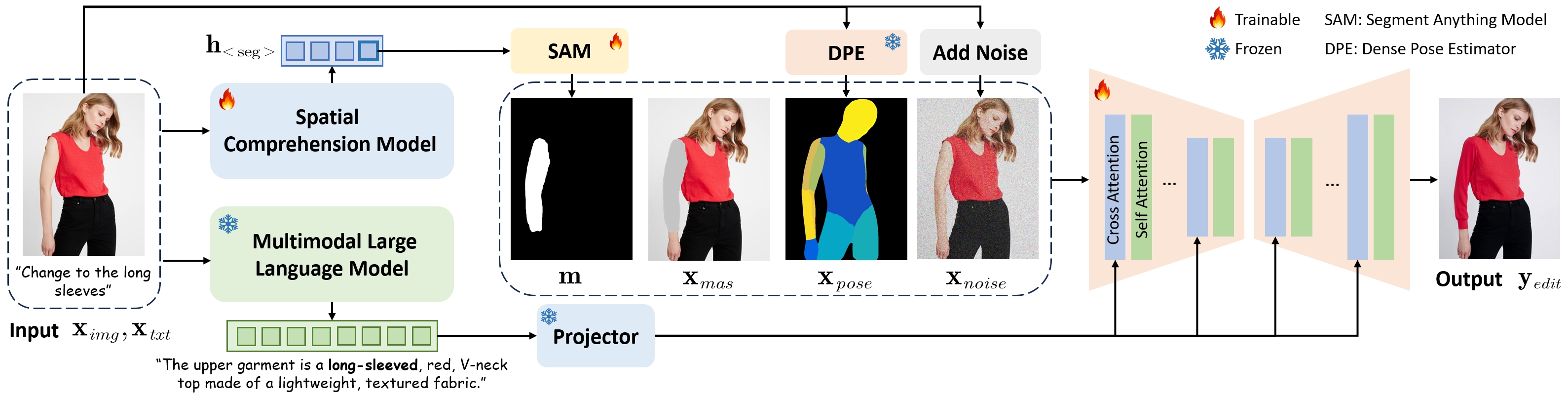} 
    \vspace{-14pt}
    \caption{\textbf{The pipeline of localized editing.} During the inference process, the image $\mathbf{x}_{img}$ and text instruction $\mathbf{x}_{txt}$ are fed into the multi-modal large language model to obtain a detailed prompt. This prompt is first mapped into a generative space and provides semantic guidance through cross-attention mechanisms. Simultaneously, $\mathbf{x}_{img}$ and $\mathbf{x}_{txt}$ are processed by the spatial comprehension model, and the last-layer embedding for the \texttt{<seg>} token $\mathbf{h}_{\texttt{<seg>}}$ serves as a prompt for SAM to derive a mask $\mathbf{m}$ indicating the local regions that need editing. This mask is concatenated with masked image, dense human pose image $\mathbf{x}_{pose}$ encoded by DPE and the noised image $\mathbf{x}_{noise}$, then input into the diffusion model $\mathcal{G}_{txt}$ to offer spatial guidance. The final output is an edited image $\mathbf{y}_{edit}$ that aligns with the provided instructions.}
    \vspace{-14pt}
\label{fig-local} 
\end{figure*}

\section{Method}

\subsection{Overview of TalkFashion}

The overall framework of TalkFashion is shown in Fig. \ref{fig-overview}. TalkFashion consists of four parts: an LLM, an invocation module, a set of try-on models, and an in-shop clothes database. First, the user uploads an image $\mathbf{x}_{img}$ and inputs a text instruction $\mathbf{x}_{txt}$. Then, the LLM analyzes the user's instruction under a predefined prompt template $\mathcal{P}$ and outputs a fixed-format response. We refer to existing prompt engineering work and design the prompt template as follows: (1) {\textit{Prefix Prompt}}, which is mainly to set the role of LLM and tell it what it needs to do; (2) {\textit{Functions Information}}, including the name and description of the available functions; (3) {\textit{Fixed Output Format}}, which specifies the output format required from the LLM for easier parsing and subsequent function invocation; (4) {\textit{Few-Shot Examples}}, which help the LLM understand the task better by providing a few examples, further reinforcing the output format. See \textit{Supplementary Material} for more details. The invocation module parses the response to get the item $i$ of clothes, the details $d$ to be modified, and the specific function $\mathcal{F}$ to be called, including full outfit change or localized editing. Thus, the entire invocation can be represented as: 
\vspace{-2pt}
\begin{equation}
    \label{eq-llm}
    \left\{\mathcal{F}, i, d\right\} = \operatorname{LLM}(\mathbf{x}_{txt}, \mathcal{P}).
\end{equation}


For full outfit change, we first obtain the masked model image with the upper body, lower body, or full body obscured based on the clothing item to be changed, represented as $\mathbf{x}_{mas}^{i}$. Then, the try-on details $d$ are matched with items in the in-shop clothes database $\mathcal{D}$, recording the highest match score $\operatorname{MS}(d, \mathcal{D})$. If it is above a predefined threshold $\tau$, a garment image $\mathbf{x}_{mat}$ most fitting to the user's requirements has been found. Consequently, the image-based try-on model $\mathcal{G}_{img}$ can be directly executed; otherwise, the text-based try-on model $\mathcal{G}_{txt}$ is used. The formal definition is shown as follows:
\vspace{-2pt}
\begin{equation}
    \mathbf{y}_{edit} =
\begin{cases}
\label{eq-full}
 \mathcal{G}_{img}(\mathbf{x}_{mas}^i, \mathbf{x}_{mat}) & \text{if } \operatorname{MS}(d, \mathcal{D}) >= \tau, \\
    \mathcal{G}_{txt}(\mathbf{x}_{mas}^i, d) & \text{else},
\end{cases}
\end{equation}
where $\mathcal{G}_{img}$ is an image-based try-on network, $\mathcal{G}_{txt}$ is a text-based editing network. For localized editings, a text-based try-on model is also utilized, with the distinction that it requires a local region mask derived from user instructions, represented as $\mathbf{x}_{mas}^{d}$. Thus, the formal definition is shown as follows:
\vspace{-2pt}
\begin{equation}
\label{eq-local}
 \mathbf{y}_{edit} = \mathcal{G}_{local}(\mathbf{x}_{mas}^{d}, d),
\end{equation}
where $\mathcal{G}_{local}$ is an adaptive localization and fine-grained editing network. Finally, the output image $\mathbf{y}_{edit}$ of the generation model $\mathcal{G}$ and the reply $\mathcal{A}$ are returned to the user.



\subsection{Full Outfit Change}

Directly generating clothing images from natural language descriptions often results in issues such as insufficient detail and limited elements. Current image-based try-on methods have shown excellent results. Therefore, we propose the pipeline of full outfit change to unify image-based and text-based methods with text. First, we use an off-the-shelf human parsing model \cite{humanparsing} to obtain a model image masked according to the clothing item to be changed. Next, we calculate the matching score between the text instructions and the garment images to determine which try-on model to use. The clothing images from the in-shop clothes database and the text instructions are input into the image encoder and the text encoder, respectively, to obtain the image embedding and text embedding. The cosine similarity between the embedding vectors is used as the matching score. If the score exceeds the threshold, we proceed with an image-based try-on model. The clothing image is fed into a separate branch to extract visual features of the garment, which are then fused with the features of the masked model image, generating a realistic image of the model wearing the selected clothing. Otherwise, we directly input the text into the main try-on network, providing semantic information to guide the filling in of the image.

\subsection{Localized Editing}


Instruction-based editing methods allow users to perform complex image edits with simple text instructions, making image editing more accessible. However, these methods often affect areas that don’t need changes. In contrast, local repainting with masks focuses edits on specific regions, avoiding unnecessary modifications but requires manual mask creation, which is less user-friendly. To address these challenges, especially in fashion domain, we propose a new solution. Our approach automatically generates masks based on user text instructions, enabling precise edits to specific regions without manual input. This ensures more natural and accurate results while simplifying the user experience.

As shown in Fig. \ref{fig-local}, the proposed localized modification network consists of three main components: the main diffusion model $\mathcal{G}_{edit}$ for generation, the spatial comprehension model $\text{SCM}$ for segmentation, and the multi-modal large language model $\text{MLLM}$ for semantic comprehension. On one side, the text instruction $\mathbf{x}_{txt}$ and input image $\mathbf{x}_{img}$ are input into $\text{MLLM}$. This model generates a detailed description of the clothing content that aligns with the user's instructions. This description is projected into the generative space to provide semantic guidance for the subsequent editing process:
\vspace{-2pt}
\begin{equation}
\label{eq-mllm}
\mathbf{E}_{guide}=\mathcal{G}_{proj}\left( \text{MLLM}\left( \mathbf{x}_{img},\mathbf{x}_{txt} \right) \right). 
\end{equation}

\begin{table*}[!t]
\centering
\caption{\textbf{Quantitative comparison with other methods.} We evaluate PSNR, SSIM, and LPIPS for low-level similarity, CLIP image similarity
score (CLIP-Score) for high-level semantic similarity, and FID score for image fidelity. We compare the metrics under both full outfit change and localized editing on the VITON-HD dataset. The best and second-best results are demonstrated in \textbf{bold} and \underline{underlined}, respectively. }
\renewcommand{\arraystretch}{1.2}
\resizebox{0.92\textwidth}{!}{
\begin{tabular}{cl|c c c c c|c c}
\shline
\multicolumn{2}{c|}{\multirow{2}{*}{\textbf{Methods}}} & \multicolumn{5}{c|}{\textbf{Full Outfit Change}} & \multicolumn{2}{c}{\textbf{Localized editing}} \\ \cline{3-9}
& & PSNR↑ & SSIM↑ & LPIPS↓ & FID↓ & CLIP-Score↑ & FID↓ & CLIP-Score↑ \\ \hline
\multicolumn{1}{c|}{\multirow{4}{*}{Mask-required}} & SDXL & \underline{40.243} & \underline{0.994} & \textbf{0.003} & 34.992 & 28.075 & 16.544 & 26.588 \\ \cline{2-9} 
\multicolumn{1}{c|}{} & ControlNet & 37.695 & 0.968 & 0.027 & 51.007 & 27.440 & 38.718 & 26.549 \\ \cline{2-9} 
\multicolumn{1}{c|}{} & UltraEdit & 37.466 & 0.987 & 0.007 & 38.260 & \underline{28.215} & 19.142 & \underline{27.092} \\ \cline{2-9} 
\multicolumn{1}{c|}{} & MagicQuill & 37.107 & \underline{0.994} & \underline{0.004} & \underline{28.926} & 27.993 & \textbf{8.297} & 26.897 \\ \cline{1-9} 
\multicolumn{1}{c|}{\multirow{2}{*}{Mask-free}} & IP2P &  - &  - & -  &  - & - & 33.196 & 23.470 \\ \cline{2-9} 
\multicolumn{1}{c|}{} & Ours & \cellcolor[HTML]{EFEFEF}\textbf{41.854} & \cellcolor[HTML]{EFEFEF}\textbf{0.995} & \cellcolor[HTML]{EFEFEF}\textbf{0.003} & \cellcolor[HTML]{EFEFEF}\textbf{28.779} & \cellcolor[HTML]{EFEFEF}\textbf{28.492} & \cellcolor[HTML]{EFEFEF}\underline{9.614} & \cellcolor[HTML]{EFEFEF}\textbf{27.236} \\ 
\shline
\end{tabular}}
\vspace{-4pt}
\label{tab-comp}
\end{table*}

On the other side, $\mathbf{x}_{txt}$ and $\mathbf{x}_{img}$ are fed into SCM. The last-layer embedding $\mathbf{h}_{\texttt{<seg>}}$ is extracted to the segment anything model $\text{SAM}$ \cite{sam} to obtain a segmentation mask $\mathbf{m}=\text{SAM}(\mathbf{h}_{<\text{seg}>})$ for the target region. This mask precisely delineates the area that needs modification. The segmentation mask, masked image $\mathbf{x}_{mas}$, noised image $\mathbf{x}_{noise}$, and dense human pose image $\mathbf{x}_{pose}$ encoded by the dense pose estimator\cite{densepose} $\text{DPE}$ are concatenated and fed into the diffusion model. The output image $\mathbf{y}_{edit}$ can be written as:
\vspace{-2pt}
\begin{equation}
\label{eq-local-edit}
 \mathbf{y}_{edit}=\mathcal{G}_{edit}(\mathbf{m}, \mathbf{x}_{pose}, \mathbf{x}_{noise}, \mathbf{x}_{mas} |\mathbf{E}_{guide}).
\end{equation}

The prompt provides semantic guidance, while the segmentation mask and dense human pose image offer spatial guidance. This combination ensures that the model focuses its editing efforts on the correct region and maintains the structural integrity of the image. The final output image not only accurately reflects the user's instructions, but also achieves a natural and harmonious visual effect.

\section{Experiments}

\begin{figure}[!t]
    \centering
    \includegraphics[width=1\linewidth]{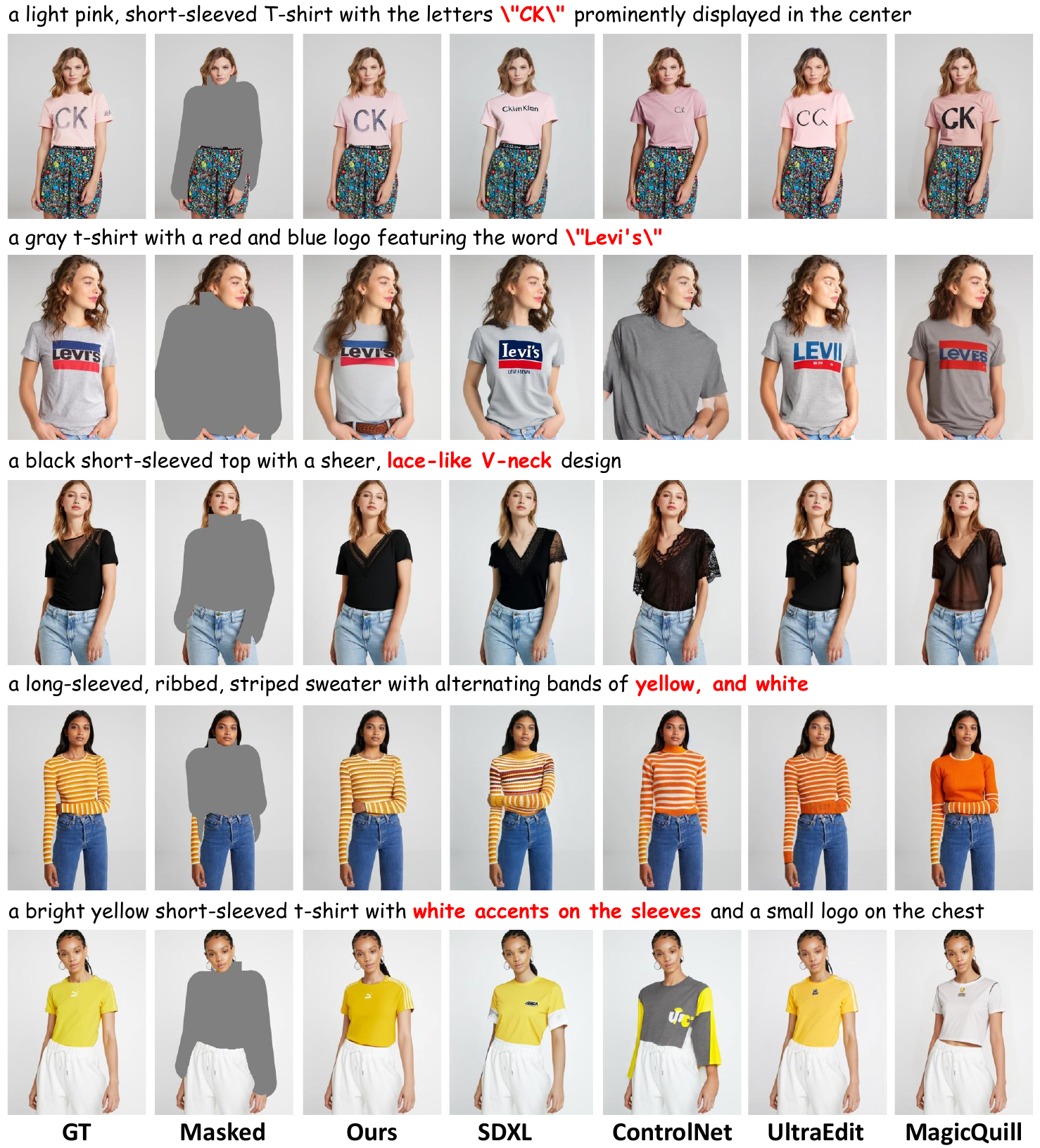} 
    \vspace{-14pt}
    \caption{\textbf{Qualitative comparisons on full outfit change.} Our method is superior to other methods in generating realistic images and ensuring semantic consistency on VITON-HD.}
    \vspace{-14pt}
\label{fig-show-full} 
\end{figure}

\begin{figure*}[!t]
    \centering
    \includegraphics[width=1\linewidth]{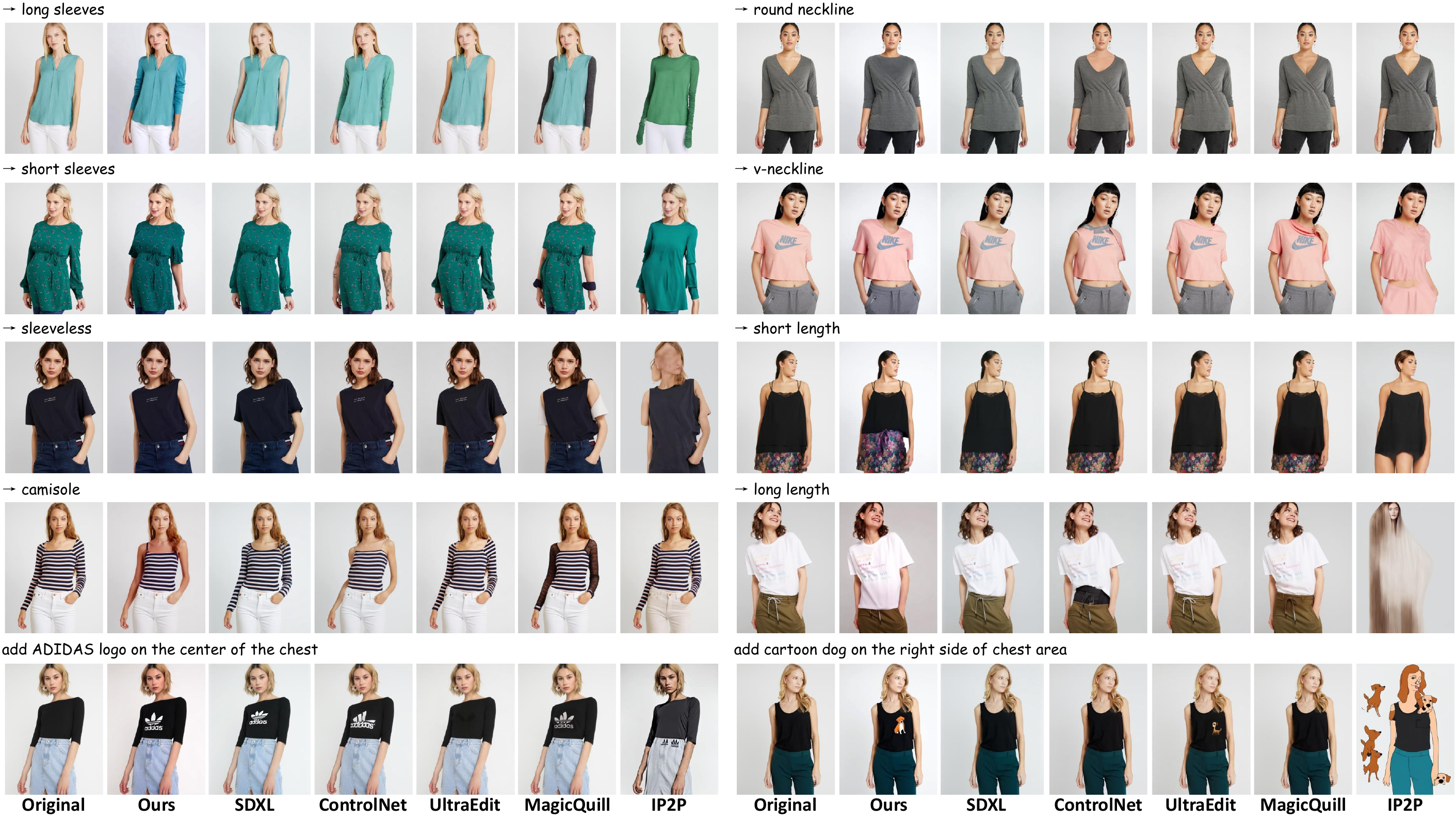} 
    \vspace{-14pt}
    \caption{\textbf{Qualitative comparisons on localized editing.} We see that the editing effect of other methods is not obvious, and it is easy to make mistakes in color and detail. Our method understands the instructions well and guarantees the fidelity.}
    \vspace{-8pt}
\label{fig-show-local} 
\end{figure*}

\subsection{Experimental Setup}

\textbf{Model Selection.} We use Qwen2-7B-Instruct~\cite{qwen2} to perform function invocation. For full outfit change, BLIP2~\cite{blip2} is used to encode the text  and image. For localized editing, Qwen2-VL-7B~\cite{qwen2vl} is utilized to serve as the prompt refiner. The diffusion model is SDXL\cite{sdxl}. 

\textbf{Training Details.} The in-shop clothes database primarily consists of the publicly available fashion dataset VITON-HD~\cite{vitonhd}. The entire network is trained in two stages on a single NVIDIA A100 80G GPU. In the first stage, we select 5,000 model images from the VITON-HD training set for manual annotation (see \textit{Supplementary Material}), involving 10 areas such as different sleeve lengths and neckline styles. We then generate relevant attribute modification instructions through the GPT-4o to obtain 5,000 instruction-mask pairs to fine-tune the segmentation modules, with a learning rate of 3e-4, for 10 epochs, and a batch size of 4. In the second stage, we fine-tune the U-Net of the diffusion model, training for 130 epochs with a learning rate of 5e-6 and a batch size of 16.

\textbf{Evaluation Metrics.} PSNR, SSIM, and LPIPS\cite{lpips} are used to measure the fidelity between the reconstructed image and the original image, thereby showing the robust performance of the model given a description of the original clothes. Meanwhile, CLIP\cite{clip} image similarity score (CLIP-Score) and FID\cite{fid} are used to measure the consistency with text instructions and the quality of generated images.


\subsection{Quantitative and Qualitative Results}

\textbf{Quantitative Comparison.} We compare our method with InstructPix2Pix (IP2P)\cite{instructpix2pix}, SDXL-Inpainting (SDXL)\cite{sdxl}, ControlNet-Inpainting (ControlNet)\cite{controlnet}, UltraEdit\cite{ultraedit}, and MagicQuill\cite{magicquill}, among which IP2P can only be edited by text, so only localized editing task is compared. The results are presented in Tab. \ref{tab-comp}. Our method achieves optimal performance across nearly all metrics on both tasks, demonstrating the effectiveness of our invocation strategy and model architecture. 


\textbf{Qualitative Comparison.} Fig. \ref{fig-show-full} shows the qualitative results of ours compared to other methods on full outfit change. We see that ControlNet tends to generate grotesque images of the human body. UltraEdit and MagicQuill exhibit low sensitivity to color perception in text. SDXL lacks refinement in the details it generates. Our results not only best meet the semantic requirements but also offer the most natural visual appearance. Fig. \ref{fig-show-local} shows the visual comparisons of ours with other methods on localized editing. In order to make a fair comparison, we use fixed-format instructions that involve multiple different clothing attributes. We can see that our method has superior editing capabilities. Other methods either do not make any changes to the image or have obvious traces of editing. IP2P, as a text-only controlled editing method, easily introduces modifications to the entire image, making it difficult to accurately edit specific regions or attributes. More qualitative results can be found in the \textit{Supplementary Material}.

\textbf{Effect of MLLM.} We conduct an ablation study on the effect of MLLM. For comparison, we train another model with attribute annotations for the VITON-HD dataset provided by IDM-VTON\cite{idmvton}. Each clothing image is associated with a series of fashion attribute tags. The quantitative results, as presented in Tab. \ref{tab-aba}, show improvements in both FID and CLIP-Score metrics. These results confirm that the introduction of MLLM significantly enhances the localized editing model's semantic understanding and generation capabilities.

\begin{table}[t]
\centering
\caption{Abalation on MLLM. We ablate the effect of using MLLM as a prompt refiner in localized editing by evaluating FID and CLIP image similarity scores.}
\renewcommand{\arraystretch}{1.2}
\setlength{\tabcolsep}{5mm}
\resizebox{0.7\linewidth}{!}{
\begin{tabular}{c|cc}
\shline
MLLM & FID↓ & CLIP-Score↑ \\ \hline
\ding{56} & 13.208 & 26.628 \\
\ding{52} & 9.614 & 27.236 \\ \shline
\end{tabular}}
\vspace{-24pt}
\label{tab-aba}
\end{table}

\section{Conclusion}

In this paper, we introduce TalkFashion, an intelligent virtual try-on assistant designed for full outfit change and localized editing. We have carefully designed the invocation strategy, allowing users to input text instructions without requiring any additional operations. For full outfit change, we incorporate the concept of image-text matching, unifying image-based and text-based virtual try-on methods through text. For localized editing, we propose a new model that leverages multimodal large models to automatically locate the regions to be edited, achieving specific attribute modifications. Experimental results demonstrate that the images generated by our method are more natural and semantically consistent.

\textbf{Limitations.} The choice of different image-based and text-based try-on models can influence the predefined matching score thresholds used in full outfit change. Meanwhile, while the network framework for localized editing has proven effective, incorporating image references could make the editable attributes more flexible and diverse, which is worth exploring.

\bibliographystyle{IEEEbib}
\bibliography{icme2025references}

\end{document}